%% file: bmvc_final.tex
\documentclass{bmvc2k}

\title{VOR-Bench: A Human Perception-Driven Benchmark for Video Object Removal}

\newcommand{\corrauth}{\ensuremath{\dagger}}
\addauthor{Haonan Huang}{}{1,*}
\addauthor{Tianrui Qiu}{}{2,3,*}
\addauthor{Xianghao Zang}{}{2}
\addauthor{Yinan Du}{}{1}
\addauthor{Zhixiang He}{}{2}
\addauthor{Chi Zhang}{}{2}
\addauthor{Hao Sun}{}{2}
\addauthor{Zhongjiang He}{}{2}
\addauthor{Tianwei Cao}{}{1}
\addauthor{Xuchong Zhang}{}{3}
\addauthor{Hongbin Sun}{}{3}
\addauthor{Kongming Liang}{}{1,\corrauth}
\addauthor{Zhanyu Ma}{}{1}

\addinstitution{
 Beijing University of\\
 Posts and Telecommunications
}
\addinstitution{
 China Telecom\\ 
 Artificial Intelligence\\
 Technology Co. Ltd
}
\addinstitution{
 State Key Laboratory of \\
 Human-Machine Hybrid Augmented \\
 Intelligence, and Institute of \\
 Artificial Intelligence and Robotics, \\
 Xi'an Jiaotong University
}

\runninghead{Haonan Huang}{VOR-Bench: A Human Perception-Driven Benchmark for VOR}

\usepackage{graphicx}
\usepackage{booktabs}
\usepackage{wrapfig}
\usepackage{pifont}
\usepackage{array}
\usepackage{multirow}
\usepackage{tabularx}
\usepackage{stfloats}
\usepackage{threeparttable}
\usepackage{tikz}
\usepackage[table,xcdraw]{xcolor}
\usepackage{float}
\usepackage{capt-of}
\usepackage[capitalize]{cleveref}
\begin{document}

\maketitle

\begingroup
\makeatletter
\renewcommand{\thefootnote}{}
\renewcommand{\@makefntext}[1]{\noindent #1}
\footnotetext{%
\textsuperscript{*} Equal contribution.
\quad
\textsuperscript{\(\dagger\)} Corresponding author.%
}
\makeatother
\endgroup

\begin{abstract}
Despite its crucial role in video object removal (VOR), existing evaluation paradigms face two critical limitations: questionable references and a misalignment between traditional metrics and human preference. To address these challenges, we introduce VOR-Bench, which advances VOR evaluation through three integrated components. First, we present the VOR Dataset (VORD), the first benchmark dataset providing both paired edited videos and graffiti masks. Its unique strength lies in a diverse data spectrum, which encompasses model-generated, tool-rendered, and camera-captured data, ensuring robust assessment across real-world scenarios. Second, we develop rMPAF, a realistic Motion-capable Paired-video Acquisition Framework. By combining the strengths of image-based object removal and fine-tuned video generation models, rMPAF automatically generates realistic, motion-coherent paired videos. Finally, we propose three evaluation dimensions and introduce VOR-MDSM, the first perception-driven VLM-based scoring model specifically designed for mask-guided VOR. It bridges the gap between arithmetic metrics and human perception by covering the essential visual attributes and matching nuanced human judgment. Extensive experiments demonstrate that VOR-Bench yields evaluation results that align closely with human perception, achieving a remarkable correlation ($\rho>0.9$) with subjective assessments. We will release VOR-Bench along with its documentation to ensure full reproducibility.
\end{abstract}

\input{sec/1_intro}
\input{sec/2_related}
\input{sec/3_method}
\input{sec/4_experiment}

\input{sec/5_conclusion}

\section*{Acknowledgements}
This work was supported by the National Nature Science Foundation of China (Grant 62476029, 62225601, 62506043), and sponsored by Beijing Nova Program and the Beijing Key Laboratory of Multimodal Data Intelligent Perception and Governance.

\bibliography{egbib}
\input{sec/X_suppl}

\end{document}

%% file: sec/1_intro.tex
\section{Introduction}
\label{sec:intro}

Recent years have witnessed rapid advancements in the field of video editing~\cite{ceylan2023pix2video,liu2024video,wu2024towards,Jiang_2025_ICCV,chen2025disco,ju2025editverse,weng2025vires}. As a crucial subtask within video editing, video object removal (VOR) aims to eliminate objects from video sequences and generate visually plausible content that is consistent with the surrounding background context. 
Given its significant challenges, this task has garnered extensive attention, catalyzing a proliferation of innovative methodologies that have evolved from early 3D convolutional neural networks~\cite{chang2019free,hu2020proposal,wang2019video} to approaches leveraging optical flow and transformers~\cite{zhang2022flow,zhou2023propainter}, and, most recently, to emerging diffusion models~\cite{li2025diffueraser,wu2025ditpainter,liu2025eraserdit,miao2026rose,zi2025minimax,yang2025mtv,samuel2025omnimattezero,fu2026effecterase,motamed2026void}.


\begin{figure}[t]
\centering
  \begin{minipage}[t]{0.44\linewidth}
  \vspace{-53pt}
    \centering
    \includegraphics[width = \linewidth]{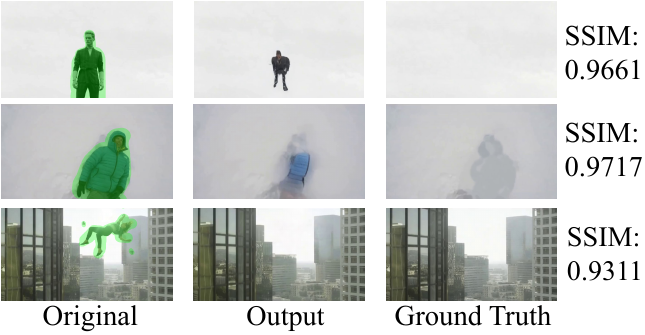}
    \vspace{-5pt}
    \captionof{figure}{The limitation of traditional metrics in estimating unexpected content generation. The first two rows yield higher SSIM scores since global averaging dilutes localized defects.}
    \label{fig:old_metrics}
  \end{minipage}\hfill
  \begin{minipage}[t]{0.52\linewidth}
    \centering
    \setlength{\tabcolsep}{4pt}
    \scalebox{0.68}{
    \begin{threeparttable}
    \begin{tabular}{ccccccc}
    \toprule
    \multirow{2}{*}{\centering Dataset} & \multirow{2}{*}{\centering Size} & \multicolumn{1}{c}{Paired-E } & \multicolumn{1}{c}{Graffiti } & \multicolumn{3}{c}{Video Types} \\
    \cmidrule(lr){5-7}  
    &               &       Videos       &    Masks      & M-G & C-C & T-R \\
    \midrule
    DAVIS~\cite{pont20172017} & 90   &  &       &  & \ding{51} &  \\
    YouTube-VOS~\cite{xu2018youtube}   & 508  &  &       &  & \ding{51} &  \\
    ROSE-Bench~\cite{miao2026rose}    & 60   & \ding{51} &       &  & & \ding{51} \\
    Movies~\cite{lin2023omnimatterf}         & 5    & \ding{51} &       &  &  & \ding{51} \\
    HQVI~\cite{cho2025elevating}          & 10   & \ding{51} &      & \ding{51} &  &  \\
    VOR-Eval~\cite{fu2026effecterase}          & 43   & \ding{51} &      &      & \ding{51} &       \\
    VOR-Wild~\cite{fu2026effecterase}          & 195   &      &      &      & \ding{51} &  \\
    \midrule
    VORD        & 150  & \ding{51} & \ding{51}      & \ding{51} & \ding{51} & \ding{51} \\
    VORD-L        & 300  & \ding{51} & \ding{51}      & \ding{51} & \ding{51} & \ding{51} \\
    \bottomrule
  \end{tabular}
  \begin{tablenotes}
        \small
        \item[*] Paired-E: paired edited; \quad M-G: model-generated; \\ C-C: camera-captured; \quad T-R: tool-rendered. 
      \end{tablenotes}
    \end{threeparttable}
  }
  \vspace{5pt}
  \captionof{table}{A comparative analysis of existing datasets for video object removal and our proposed VORD and larger version VORD-L.}
  \label{tab:1}
  \end{minipage}
  \vspace{-2mm}
\end{figure}
Nevertheless, the notable progress in video removal has posed significant challenges in its comprehensive and accurate evaluation. These challenges stem from two critical aspects: datasets and evaluation metrics. On the data front, existing benchmarks lack paired edited videos, precluding direct comparisons with model removal outputs, leading to questionable evaluation references. Furthermore, they employ precise masks generated by segmentation models, which stand in contrast to the masks provided by users in real-world applications. Simultaneously, existing datasets suffer from homogeneity, being dominated by either tool-rendered data or camera-captured data, with limited diversity in scenes and objects. 

In terms of metrics, existing evaluation paradigms typically rely on traditional image-level quality metrics such as Peak Signal-to-Noise Ratio (PSNR)~\cite{hore2010image}, Structural Similarity (SSIM)~\cite{wang2004image}, and Learned Perceptual Image Patch Similarity (LPIPS)~\cite{zhang2018unreasonable}. These metrics require paired edited videos to perform calculations against a ground-truth reference. However, these metrics measure pixel-level or structural fidelity rather than semantic meaning, resulting in significant divergence from human evaluation. Meanwhile, as exemplified in \cref{fig:old_metrics}, the first two outputs contain visible artifacts yet yield higher SSIM scores. This is because its global averaging logic causes localized defects within the mask area to be diluted by the large, defect-free background. Consequently, the models can obtain high scores by optimizing background preservation while neglecting the actual quality of object removal. Even when restricted to non-masked areas for measuring background preservation, these metrics treat images as mere collections of independent pixels rather than coherent and structured wholes~\cite{chandrasekar2024remove}. This flaw prevents it from accurately capturing changes in perception. Therefore, such limitations highlight the need for a more perceptually convincing evaluation benchmark for the VOR field.

\begin{wrapfigure}{r}{0.58\linewidth} 
  \centering
  \includegraphics[width=\linewidth,keepaspectratio]{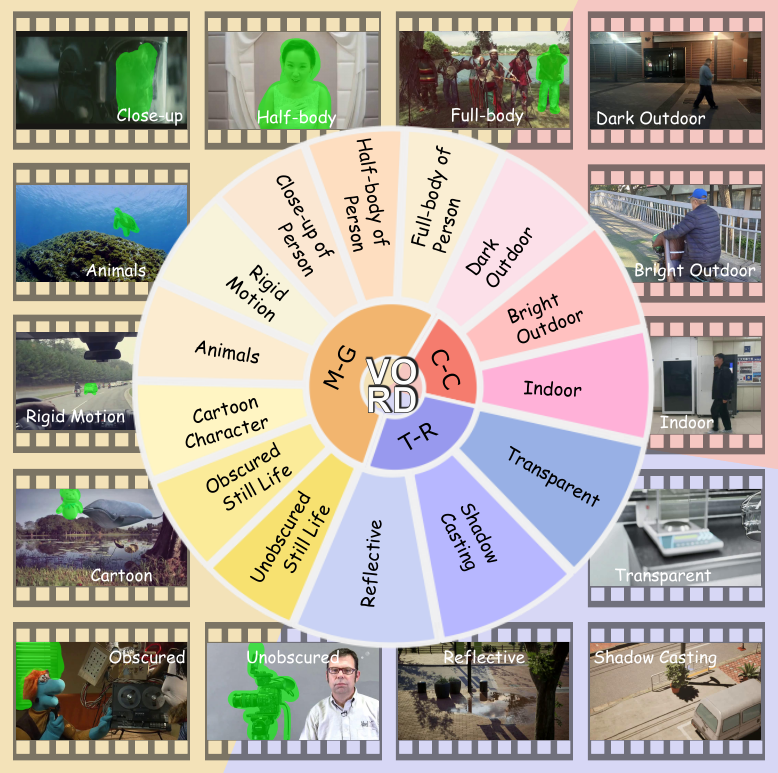}
  \caption{Diverse categories of VORD, with representative visual examples. "M-G": model-generated data; "C-C": camera-captured data; "T-R": tool-rendered data.}
  \label{fig:dataclassi}
\end{wrapfigure}

To remedy the data limitations, we introduce VORD, the first VOR evaluation dataset that uniquely features both paired edited videos and graffiti masks. VORD comprises 150 paired video sequences, a scale determined through careful consideration to balance evaluation reliability and computational cost. We further introduce a larger-scale version, VORD-L, to provide an option for broader evaluation and to serve as a supplementary benchmark for comparison; detailed ablation experiments are presented in \cref{sec:mainexperiment}. As illustrated in \cref{fig:dataclassi}, VORD covers three data types: model-generated, camera-captured, and tool-rendered. It offers a fine-grained categorization across 8 removal object types, 3 real-world scenarios, and 3 object association effects. This dataset also encompasses varying video lengths and frame rates, thereby enhancing its complexity and practicality. As shown in \cref{tab:1}, VORD distinguishes itself from existing datasets by providing not only reliable paired edited videos but also user-realistic graffiti masks. The masks for the first frames are manually painted, while a segmentation model generates the remaining masks. This dataset not only provides diverse data categories but also allows for assessing models' ability to precisely segment target objects. 

To address the critical lack of paired edited videos in existing evaluation sets, we present a realistic Motion-capable Paired-video Acquisition Framework, rMPAF. Acquiring such pairs is highly challenging, as real-world videos inherently lack removal videos, while static-camera or rendered videos suffer from limited scene diversity and restricted motion patterns. Building upon an image removal model and a trained video generation model, our framework generates high-quality paired edited videos that highly simulate real-world physical motion and maintain alignment in non-edited regions. These videos form the foundational component of our dataset, enabling precise and comprehensive model evaluation.

To supplement the current field of VOR evaluation, we innovatively propose three evaluation dimensions that comprehensively cover the core aspects of the task: object removal completeness, mask-background consistency, and logical plausibility. To quantitatively evaluate these dimensions with close alignment to human perception, we introduce VOR-MDSM, a novel VLM-based multidimensional scoring model. Built upon the strong multimodal understanding of the VLM and supervised by human ratings, VOR-MDSM formulates VOR evaluation as a question-answering task. By leveraging rich temporal-spatial video information, it outputs perceptually aligned scores across the three aforementioned dimensions, enabling a highly reliable assessment of edited videos.
Extensive experiments demonstrate that VOR-MDSM significantly complements existing evaluation methodologies, exhibiting a stronger correlation with human perceptual judgments. Our contributions can be summarized as follows:
\begin{itemize}
\item We propose VORD, the first VOR quality evaluation dataset featuring both paired edited videos and graffiti masks, characterized by diverse data types and varied video lengths. This dataset better aligns with practical application scenarios, providing comprehensive data categories for the VOR evaluation.
\item We design a novel data framework rMPAF to address the lack of paired edited videos and the homogeneity of scenarios and motion patterns in existing datasets. This framework provides a viable solution for constructing high-quality video pairs.
\item We present VOR-MDSM, a human perception-driven video removal quality scoring model based on VLM. Through deliberate designs and high-quality data, it provides a comprehensive evaluation framework encompassing three dimensions. Extensive experiments validate its superior alignment with human preference.
\end{itemize}


%% file: sec/2_related.tex
\section{Related Work}
\label{sec:formatting}

\subsection{Video Object Removal Models}

The VOR task aims to seamlessly fill masked regions with content consistent with the background context. Early approaches typically relied on mechanisms such as 3D convolutional neural networks~\cite{kim2019deep,chang2019free,hu2020proposal,wang2019video}, optical flow~\cite{li2020short,zou2021progressive,xu2019deep}, and transformer~\cite{li2022towards,cai2022devit,lee2019copy,ren2022dlformer,zeng2020learning} to achieve object removal. ProPainter~\cite{zhou2023propainter} combines optical flow with the transformer, allowing for the propagation of known pixels and the generation of unknown content. In recent years, with the rapid emergence and impressive generative capabilities of diffusion models~\cite{ho2020denoising,song2020denoising,rombach2022high,dhariwal2021diffusion,lipman2022flow,ho2022video}, a new paradigm for VOR has emerged~\cite{ju2024brushnet,bian2025videopainter,samuel2025omnimattezero,wu2025ditpainter,yang2025mtv}. 
MiniMax-Remover~\cite{zi2025minimax} focuses on developing a lightweight architecture based on the Diffusion Transformer (DiT) coupled with a two-stage training strategy, enabling efficient generation. Beyond simple object removal, Casper~\cite{lee2025generative}, ROSE~\cite{miao2026rose}, and EffectErase~\cite{fu2026effecterase} tackle the more complex problem of removing objects along with their associated effects by constructing specialized paired datasets. VOID~\cite{motamed2026void} further highlights the importance of interaction-aware object removal by focusing on objects and their induced visual interactions in realistic video scenarios. 

Despite these advancements, the fundamental evaluation of VOR remains underexplored: whether the target object is truly removed, whether the background is plausibly restored, and whether the resulting video appears natural to human observers. Existing metrics are insufficient to effectively compare diverse VOR models from a human perception perspective. To address this gap, VOR-Bench provides a diverse benchmark dataset and a human perception-driven multidimensional scoring model for evaluating the core quality of VOR.


\subsection{Video Object Removal Evaluation Datasets}

The video object segmentation datasets such as DAVIS~\cite{pont20172017} and YouTube-VOS~\cite{xu2018youtube} are widely adopted for evaluating VOR models. 
However, these existing evaluation datasets lack the necessary paired edited videos for reasonable evaluation reference. This absence inherently hinders the objective assessment of removal quality. 
Although datasets such as ROSE-Bench~\cite{miao2026rose}, Movies~\cite{lin2023omnimatterf}, HQVI~\cite{cho2025elevating} and VOR-Eval~\cite{fu2026effecterase} include paired edited videos, they are restricted to a singular generation paradigm, such as pure rendering, simple mask pasting, or direct camera capture. This limitation naturally leads to homogeneous data types, which, when coupled with a limited data scale, fundamentally fail to satisfy diverse and robust evaluation demands. To address these limitations, we construct VORD with diverse data sources and design rMPAF to generate realistic model-generated paired videos. VORD comprises 150 video pairs sourced from model-generated, tool-rendered, and camera-captured data, featuring various removal objects, representative real-world scenes, and object association effects.

\subsection{Video Object Removal Assessment}
Current VOR quality assessment typically relies on generic image similarity metrics. For instance, PSNR~\cite{hore2010image} calculates the mean squared error (MSE) between real video frames and generated video frames. SSIM~\cite{wang2004image} measures the similarity across luminance, contrast, and structure. LPIPS~\cite{zhang2018unreasonable} employs deep neural networks to extract features from video frames and compute feature distances to determine similarity. However, these metrics are limited to measuring pixel-level or structural consistency and often diverge from human perceptual preferences. Moreover, their calculations strictly depend on authentic reference images. 

Due to the limitations of traditional metrics, researchers have explored evaluation methods better aligned with human perception. Studies on forgery detection highlight that accurately identifying perceptual artifacts is crucial for judging visual authenticity~\cite{gao2023self,gao2026toward}. Recent works therefore leverage VLMs for perceptual quality assessment. Video-Bench~\cite{han2025video} combines few-shot scoring with chain-of-query techniques for video generation assessment, while ImgEdit-Judge~\cite{ye2025imgedit} and TDVE-Assessor~\cite{wang2025tdve} fine-tune Qwen2.5-VL-7B~\cite{bai2025qwen2} for image editing and text-driven video editing evaluation, respectively. In VOR, VOID~\cite{motamed2026void} prompts closed-source VLMs to assess interaction-aware removal quality. Nevertheless, such prompt-based evaluation depends on proprietary models and is not explicitly optimized for human perception-aligned VOR assessment. Therefore, our VOR-Bench fills this gap by introducing a locally deployable, cost-efficient multidimensional scoring model supervised by human annotations, offering a benchmark consistent with human perception for this task.

%% file: sec/3_method.tex
\section{VOR-Bench}

\subsection{Dataset Construction}
\label{sec:3.1}
To address the data homogeneity and questionable evaluation references of existing VOR benchmarks, VORD employs four key strategies: 1) Constructing precise paired edited videos, thereby establishing a unified and fair benchmark for VOR quality assessment. 2) Acquiring diverse data types, ensuring comprehensive coverage of model-generated data, tool-rendered data, and camera-captured data. 3) Incorporating graffiti masks, which simulate the practical user interactions common in real video editing scenarios. 4) Determining an optimal dataset scale, which balances statistical significance and evaluation efficiency. VORD finally comprises 150 video pairs with 14 different visual types, and three data types in equal proportion. More visual samples have been shown in \cref{fig:dataclassi}. \\
\textbf{Model-generated data construction.} Obtaining pairs of original and removal videos in VOR tasks poses a challenge for existing evaluation systems. To address this challenge, we propose a novel data framework, rMPAF, that leverages the 2D object removal model GeoRemover~\cite{zhu2025georemover} and the video generation model Wan-2.1~\cite{wan2025wan}. Each generated video contains 41 frames at a frame rate of 15 fps. For specific details, please refer to \cref{sec:pipeline}.\\
\textbf{Camera-captured data acquisition.} To acquire camera-captured data that captures authentic object environment interaction effects, we utilized fixed cameras to capture 50 pairs of video sequences. Each collected video pair contains objects entering and exiting the frame and each video contains 150 frames at 30 fps. This design facilitates the generation of pseudo-tuples comprising an object present video and its object removed counterpart. The camera-captured data cover diverse lighting variations, but they also have limitations in capturing motion due to fixed camera positions. Our model-generated data compensates for this shortcoming.\\
\textbf{Tool-rendered data collection.} For our tool-rendered data, we considered the challenges of long duration and effects removal, and selected a total of 50 items from the open-source ROSE-Bench~\cite{miao2026rose} and Movies~\cite{lin2023omnimatterf}. The tool-rendered data encompasses three different effects: transparent, shadow casting and reflective. \cref{fig:dataclassi} shows the specific samples. \\
\textbf{Mask generation strategy.} To better accommodate irregular masks that users may provide, we manually painted a graffiti mask for the first frame of each video, while extracting segmentation masks for subsequent frames using Grounded SAM2~\cite{ravi2024sam}.\\
\textbf{Dataset scale selection.} The evaluation benchmark requires a careful trade-off among scenario diversity, statistical significance, and inference efficiency. Considering that some advanced VOR models incur substantial computational costs during inference, we deliberately select 150 video pairs as the evaluation scale. While ensuring stable and discriminative evaluation rankings, the current benchmark provides extensive coverage of practical scenarios without introducing unnecessary computational overhead.\\
\textbf{Correctness of data.} Following CineTechBench~\cite{wang2025cinetechbench}, we used MonST3R to estimate camera trajectories and computed Rotation Error (RotErr), scene-scale-normalized Translation Error (TransErr), and Camera Motion Consistency (CamMC), which measure rotational, translational, and joint camera-motion discrepancies, respectively. TransErr is not measured in metres; these metrics evaluate camera-motion consistency rather than pixel identity.
\begin{wraptable}{r}{0.57\linewidth}
  \centering
  \vspace{-0.3mm}
  \scriptsize
  \begin{threeparttable}
  \begin{tabular}{cccccc}
    \toprule
    \multirow{2}{*}{\centering Models} & \multirow{2}{*}{\centering RotError $\downarrow$} & \multicolumn{2}{c}{TransError $\downarrow$} & \multicolumn{2}{c}{CamMC $\downarrow$}  \\
    \cmidrule(lr){3-4}\cmidrule(lr){5-6}
    &  & Rel.\tnote{*}& Abs.\tnote{*} & Rel. & Abs. \\
    \midrule
    Klingv1.6 & 21.68 & 48.49 & 196.14 & 62.57 & 207.65  \\
    Wan2.1~\cite{wan2025wan} & 27.80 & 48.31 & 99.61 & 67.82 & 115.76  \\
    \midrule
    Ours & 0.40  & 23.48 & 64.05 & 23.62 & 64.16  \\
    \bottomrule
  \end{tabular}
  \begin{tablenotes}
        \scriptsize
        \item[*] Rel. represents the relative score of the metric. Abs. represents the absolute score of the metric. 
      \end{tablenotes}
    \end{threeparttable}
  \vspace{4mm}
  \caption{Proof of model-generated data correctness.}
  \label{tab:data_effect}
  \vspace{-4mm}
\end{wraptable}
As depicted in \cref{tab:data_effect}, our model-generated data achieved the lowest values for the three metrics respectively. These values are significantly lower than the corresponding metrics reported within the CineTechBench dataset. This demonstrates the correctness and reliability of our model-generated data, laying the foundation for subsequent research. For camera-captured data, due to the complex changes in real-world scenarios, we chose a fixed perspective to ensure that qualified video pairs can be obtained in adjacent time periods, thus guaranteeing background consistency. The construction process of tool-rendered data is already systematic, further details are omitted.


\begin{figure}[t]
    \centering
    \includegraphics[width=\linewidth]
    {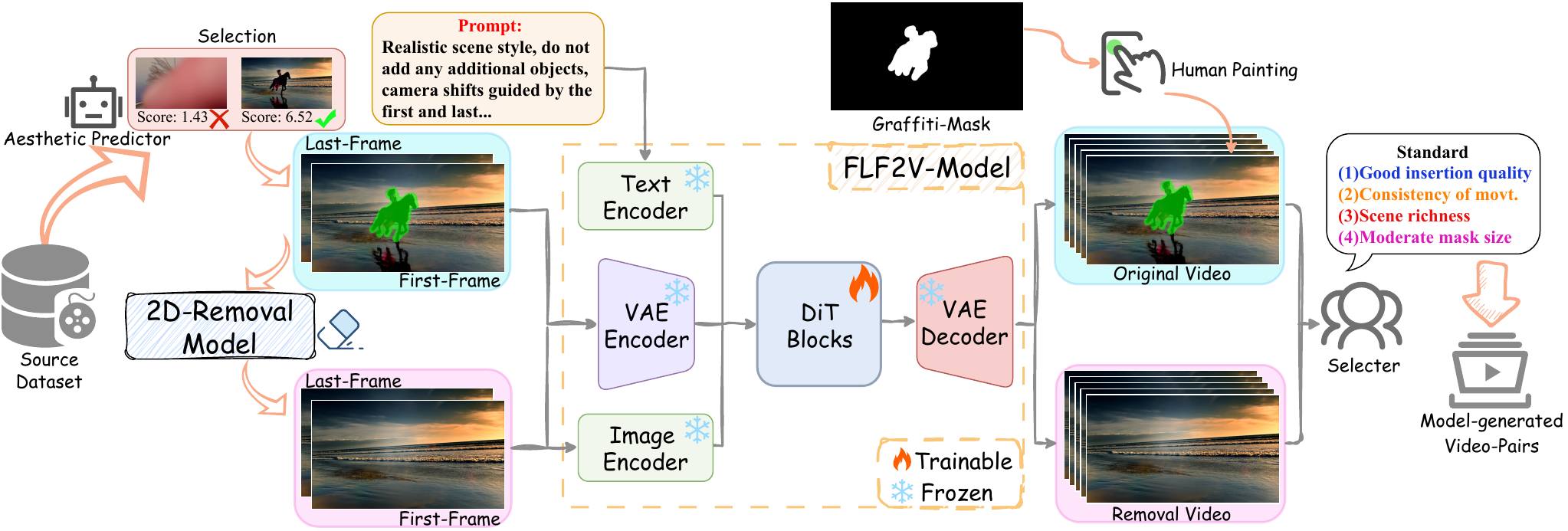}
    \caption{
        A realistic Motion-capable Paired-video Acquisition Framework, rMPAF. It comprises five stages: (1) source data collection (2) aesthetic filtering (3) first \& last frame object removal (4) frame interpolation by FLF2V-Model (5) final selection. Specifically, FLF2V-Model is used to generate intermediate frames by first frame and last frame.
    }
    \label{fig:datapipeline} 
    \vspace{-10pt} 
\end{figure}
\subsection{Realistic Motion-capable Paired-video Acquisition Framework}
\label{sec:pipeline}
To overcome the critical scarcity of paired videos with diverse scenes and natural dynamics, we present the rMPAF. As depicted in \cref{fig:datapipeline}, rMPAF is systematically divided into five stages: source data collection, aesthetic filtering, first \& last frame object removal, video generation via frame interpolation, and final selection. We utilize Vchitect-T2V-Dataverse~\cite{fan2025vchitect} as our source data, given its diverse and high-quality video content. To filter out higher-quality data, we use the open-source Aesthetic Predictor to remove data with low scores for the first frame. Subsequently, we extract the first and last frames from these source videos. These frames are then processed by the 2D-removal model GeoRemover to generate object removed first and last frames. The two pairs of first and last frames are then fed into the fine-tuned Wan-2.1-FLF2V model for frame interpolation, yielding the target video pairs. 

Specifically, we fine-tune Wan-2.1-FLF2V using low-rank adaptation (LoRA)~\cite{hu2022lora} with 100 videos mainly selected from DAVIS and HQVI. We choose them because they contain diverse real-world scenes, high-quality video content, and natural camera motion, which are important for learning realistic interpolation between the first and last frames. Such data provides a suitable basis for generating paired edited videos with realistic scene appearance and camera movement. This targeted fine-tuning aims to ensure that the interpolation quality is optimized for the specific data requirements of the VOR task, especially maintaining consistency in the removed regions. To ensure the reliability of generated pairs, the original and removed videos are not independently generated. Instead, they share identical first and last frame anchors, and the same fine-tuned FLF2V model is applied under identical settings to interpolate intermediate frames. We further manually filter out samples with temporal inconsistency, removal failures, visual artifacts, or invalid masks. 

Furthermore, compared with the existing tool-rendered data with fixed objects, the data constructed by this framework not only has a unified style but also fits real scenarios better, conforms to real motion laws without being limited to rigid objects. It also yields more diverse shot-trajectory data with lower cost than camera-captured data.

\begin{figure}[!b]
    \centering
    \includegraphics[width=0.85\linewidth]{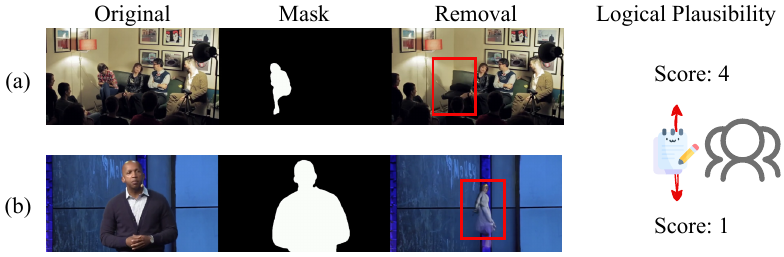}
    \caption{
         Visual examples of the logical plausibility score.
    }
    \label{fig:logical} 
\end{figure}
\subsection{Evaluation Metrics} 
Aiming to establish an evaluation framework that captures the entire visual spectrum of VOR tasks while aligning with human perception, we introduce three new evaluation dimensions: object removal completeness, mask-background consistency and logical plausibility. All three dimensions are scored from 1 to 5 using our multidimensional scoring model.\\
\textbf{Object removal completeness (ORC).} As the core objective of VOR task, object removal completeness primarily evaluates whether the masked object has been completely eliminated. The score ranges from 1 to 5, reflecting the removal proportion of the masked object, with higher scores indicating more complete elimination.\\
\textbf{Mask-background consistency (MBC).} We focus on whether the content generated within the masked region blends naturally with the surrounding background. Mask-background consistency aims to evaluate the consistency in details such as color, texture, material and patterns. It is an important measure of whether a video removal model could achieve seamless inpainting.\\
\textbf{Logical plausibility (LP).} It assesses whether the content generated by the model in the masked region is semantically related to the background and adheres to physical laws and common sense. As shown in (a) of \cref{fig:logical}, the original structure of the sofa in the removal video has been almost preserved with flaws only visible when zoomed in; (b) shows the redundant object in the masked area and the extremely poor removal effect.\\
\textbf{Temporal consistency.} Temporal consistency is considered within the video-level MBC/LP criteria rather than treated as an independent dimension. MBC penalizes temporally unstable blending in the removed region, while LP penalizes cross-frame violations of scene logic and physical plausibility. In the dimension of LP, we have treated unreasonable motion of a still object as a low-score case, directly covering a temporal-consistency failure. Thus, temporal consistency is not ignored, but incorporated into MBC/LP. We did not add a separate TC dimension because temporal failures in VOR are highly coupled with MBC/LP, which could cause duplicated penalties and annotation ambiguity.

The scoring criteria for the above dimensions will be explained in the supplementary materials (Section 6).

\begin{figure}[!t]
    \centering
    \includegraphics[width=\linewidth]{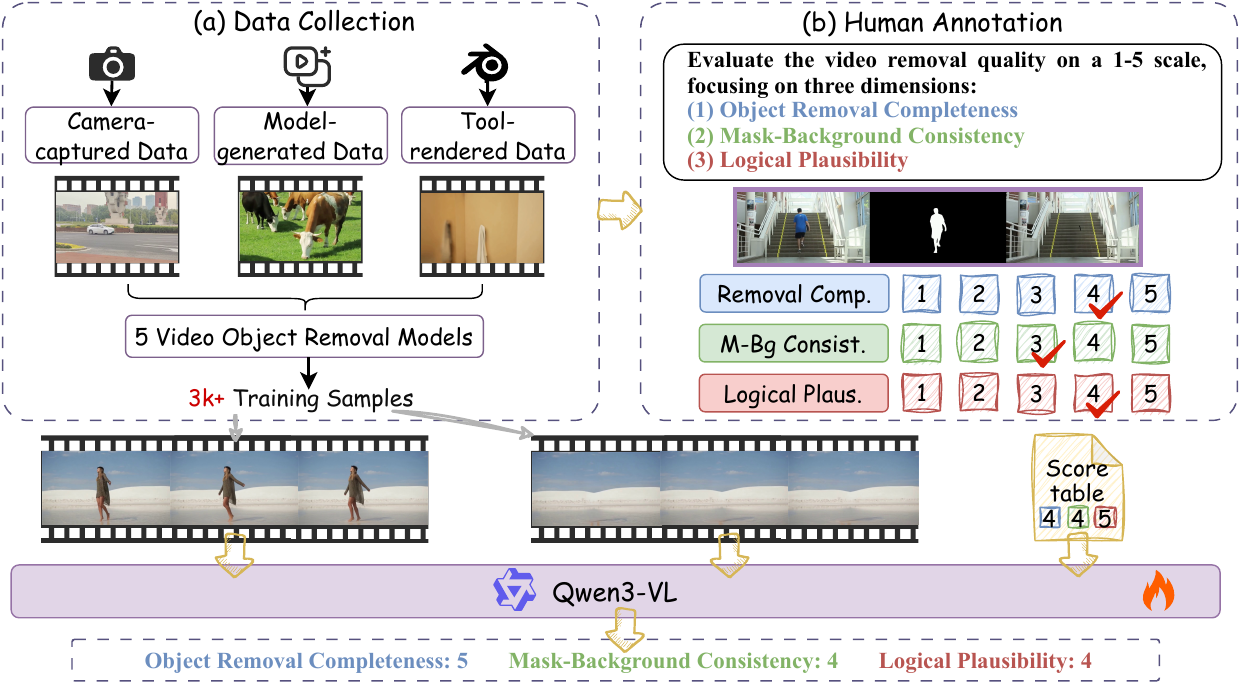}
    \caption{
         Overview of VOR-MDSM. (a) We utilize rMPAF (\cref{{sec:pipeline}}) to gather three categories of source data and employ various VOR models to yield the final training dataset;
        (b) Vision experts annotate this training data across three distinct dimensions.
        Finally, the VLM is subsequently fine-tuned using the annotated training data.
    }
    \label{fig:qwen} 
\end{figure}

\subsection{VOR-MDSM}
\textbf{Training dataset construction.} To construct our training dataset, source videos were collected using the same data categories and construction methods as described for the test dataset. To ensure a diverse and extensive training corpus, we employed five different open-source VOR models to process these source videos. This process ultimately yielded 3,299 edited video pairs. Subsequently, we recruited 15 vision experts, organized into three groups, to conduct human annotation on these video pairs. For each video pair, annotators were tasked with evaluating the quality of object removal in the generated video across three dimensions. Prior to the formal annotation process, volunteers received comprehensive training on the scoring criteria for each of the three dimensions. They were also provided with a detailed annotation guideline for reference. Each video pair received ratings from five independent annotators. The final annotation score for each pair was then determined by averaging these five ratings. These human-annotated scores are used as supervisory signals to train the VOR-MDSM, enabling it to learn human-aligned quality assessment. The detailed information of the experts will be explained in the supplementary materials (Section 6).\\
\textbf{Multidimensional scoring model.} Considering the scoring requirements for associating visual information and text semantics in multi-modal input, we have selected a VLM as the evaluator, termed VOR-MDSM. To address the current void in open-source VLM evaluators specifically designed for VOR tasks, 
we employ LoRA for parameter-efficient fine-tuning of the VLM using the previously constructed training corpus. \cref{fig:qwen} depicts the overall architecture of the VOR-MDSM. It provides a three-dimensional rating directly from a pair of original and edited videos. This evaluation paradigm assesses editing results without requiring removal ground truth. Trained under the supervision of human-annotated data, VOR-MDSM requires only video pairs to generate scores aligned with human perception. Unlike conventional image quality assessment (IQA) metrics that rely on low-level visual similarity, VOR-MDSM performs task-specific comparison between original and edited videos, focusing on object removal completeness, mask-background consistency, and logical plausibility.

Based on the comparative experiments described in \cref{sec:mainexperiment}, we select the optimal training data input format. Specifically, each training sample is formulated as \(x_i=(v_i^{orig}, v_i^{edit}, q_i)\), where \(q_i\) is the scoring instruction. The target is a structured textual answer containing three scores \(s_i=(s_i^{orc},s_i^{mbc},s_i^{lp})\in\{1,\ldots,5\}^3\), e.g., ``ORC: 3, MBC: 3, LP: 4''. VOR-MDSM is trained by supervised fine-tuning Qwen3-VL with LoRA to generate this scoring text, rather than by three-dimensional score regression. The objective is the standard token-level causal language modeling loss:
\begin{equation}
\setlength{\abovedisplayskip}{3pt}
\setlength{\belowdisplayskip}{3pt}
\mathcal{L}(\theta)=-\frac{1}{N}\sum_i\sum_{t\in\mathcal{A}_i}
\log p_\theta(y_{i,t}\mid x_i,y_{i,<t}),
\label{eq:sft_loss}
\end{equation}
where \(\theta\) denotes the trainable model parameters, \(N\) is the number of training samples, \(\mathcal{A}_i\) denotes the set of assistant-answer token positions for the \(i\)-th sample, \(y_{i,t}\) is the target token at position \(t\), and \(y_{i,<t}\) represents the preceding target tokens. 

Only assistant-answer tokens are supervised, while prompt and video tokens are ignored in the loss computation. Additionally, the perception alignment evaluation and cross-dataset ablation in \cref{sec:mainexperiment} further validate the accuracy and robustness of VOR-MDSM.\\
\textbf{Implementation details.} Using Qwen3-VL-8B~\cite{yang2025qwen3} as the backbone, we performed training for 10K steps on 4 NVIDIA A100 GPUs with LoRA rank of 64. The input resolution for the training data was 992×560, with a batch size of 2 and a learning rate of 1e-4.

%% file: sec/4_experiment.tex
\section{Experiments}
\subsection{Experimental Setup}
For comparative analysis, we choose ten open-source representative VOR models: FuseFormer~\cite{liu2021fuseformer}, FGT~\cite{zhang2022flow}, ProPainter~\cite{zhou2023propainter}, FloED~\cite{gu2024coherent}, Casper~\cite{lee2025generative}, MiniMax-Remover~\cite{zi2025minimax}, DiffuEraser~\cite{li2025diffueraser}, ROSE~\cite{miao2026rose}, EffectErase~\cite{fu2026effecterase}, VOID~\cite{motamed2026void}. This selection intentionally spans a wide range of methodologies, encompassing approaches from foundational works to recent state-of-the-art advancements in the field. We run VORD on these models, with the same input and output resolution. Then we score them by three distinct dimensions: object removal completeness, mask-background consistency, and logical plausibility, using our scoring model. 
\begin{table}[t]
  \centering
  

  \renewcommand{\arraystretch}{1.3}
  \scriptsize
  \setlength{\tabcolsep}{3.5pt} 

  \begin{tabular*}{\linewidth}{@{\extracolsep{\fill}}l c c c c c c c c}
    \toprule
    \multirow{2}{*}{Models} &
    \multirow{2}{*}{Year} &
    \multicolumn{3}{c}{Background Preservation} &
    \multicolumn{4}{c}{VOR-MDSM Score} \\
    \cmidrule(lr){3-5} \cmidrule(lr){6-9}
    & &  PSNR$\uparrow$ & SSIM$\uparrow$ & LPIPS$\downarrow$ & ORC\tnote{*} & MBC\tnote{*} & LP\tnote{*} & Avg \\
    \midrule
    FuseFormer~\cite{liu2021fuseformer}  & 2021 & 27.941& 0.890& 0.070 & 3.25 & 3.22 & 3.71 & 3.39  \\
    FGT~\cite{zhang2022flow}             & 2022 & 28.630 & 0.882 & 0.083 & 3.57 & 3.39 & 4.05 & 3.67  \\
    ProPainter~\cite{zhou2023propainter} & 2023 & 29.644 & \textbf{0.914} & \underline{0.053} & 3.31 & 3.14 & 3.69 & 3.38  \\
    FloED~\cite{gu2024coherent}          & 2024 & 27.832 & 0.891 & 0.069 & 3.20 & 3.17 & 3.61 & 3.33  \\
    Casper~\cite{lee2025generative}      & 2025 & 29.449 & 0.893 & 0.057 & 4.03 & 4.03 & 4.38 & 4.15  \\
    DiffuEraser~\cite{li2025diffueraser} & 2025 & 29.394 & 0.910 & 0.056 & 3.77 & 3.69 & 4.20 & 3.89  \\
    MiniMax-Remover~\cite{zi2025minimax} & 2025 & 28.880 & 0.898 & 0.060 & 3.99 & 3.91 & 4.37 & 4.09  \\
    ROSE~\cite{miao2026rose}             & 2025 & \underline{30.412} & 0.900 & \underline{0.053} & \textbf{4.27} & \textbf{4.11} & \textbf{4.61} & \textbf{4.33}  \\
    EffectErase~\cite{fu2026effecterase}             & 2026 & 28.723 & 0.889 & 0.064 & \underline{4.19} & 4.02 & \underline{4.59} & \underline{4.27}  \\
    VOID~\cite{motamed2026void}             & 2026 & \textbf{30.988} & \underline{0.912} & \textbf{0.045} & 4.04 & \underline{4.04} & 4.57 & 4.22  \\
    \bottomrule
  \end{tabular*}


  \vspace{4mm}
    \caption{Overall quantitative comparisons. It comprises two parts: background preservation and three VLM-evaluated dimensions. Bold indicates the best; underlined indicates the second-best.}
    \label{tab:overall}
\end{table}
\subsection{Evaluation Results}
\label{sec:experiment}

\textbf{Quantitative evaluation.} \Cref{tab:overall} compares ten representative VOR methods. ROSE achieves the best overall performance across the three VOR-MDSM dimensions, followed by EffectErase and VOID. The relatively lower scores of other methods demonstrate the capability of VOR-MDSM in distinguishing different removal qualities. Notably, mask-background consistency remains the weakest dimension across most methods, indicating that background restoration is still a major challenge for current VOR models and motivating future efforts toward improving this aspect.

Furthermore, we evaluate traditional metrics on non-masked regions to examine whether they can better reflect background preservation. However, the resulting rankings differ from both VOR-MDSM and qualitative comparisons, suggesting limited alignment with human perception. Specifically, ProPainter achieves higher traditional metric scores than ROSE, whereas ROSE obtains substantially better VOR-MDSM scores and qualitative results. This indicates that pixel-level similarity alone cannot fully capture VOR-specific quality factors, as better background preservation may not necessarily correspond to better removal quality. Therefore, we report model rankings based on VOR-MDSM rather than traditional metrics.

\begin{figure}[t]
    \centering
    \includegraphics[width=\linewidth]{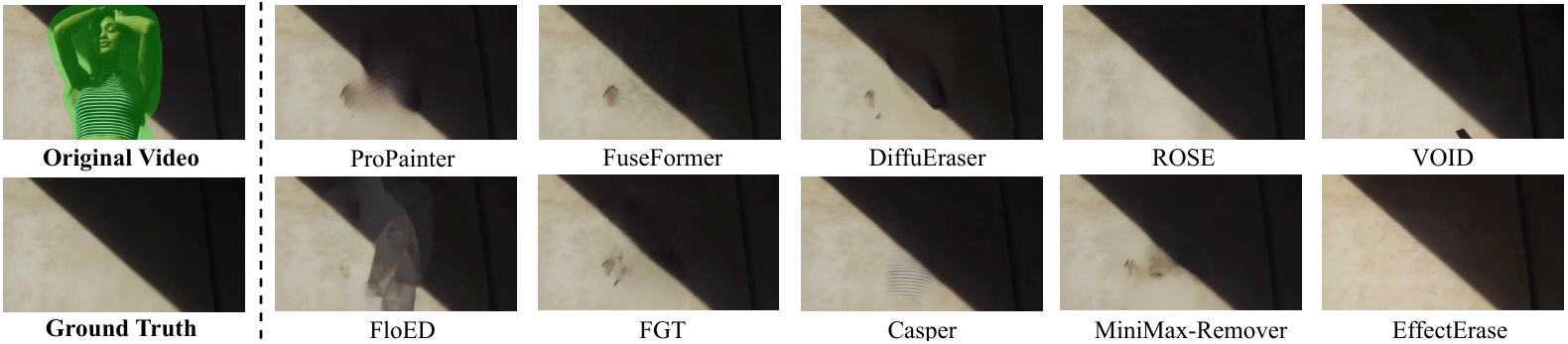}
    \caption{
         The visual results of different models. The video frames on the left depict the original video and ground-truth video.
    }
    \label{fig:visual} 
\end{figure}

\noindent\textbf{Qualitative evaluation.} \Cref{fig:visual} supports the quantitative ranking: ROSE produces the cleanest result, whereas other methods exhibit more visible artifacts or removal boundaries. Logical errors are relatively rare, consistent with the generally high LP scores in \cref{tab:overall}. Additional visual results are provided in the supplementary material (Section 8).

\subsection{Ablation Study}
\label{sec:mainexperiment}
\begin{figure}[t]
    \centering
    \includegraphics[width=\linewidth]{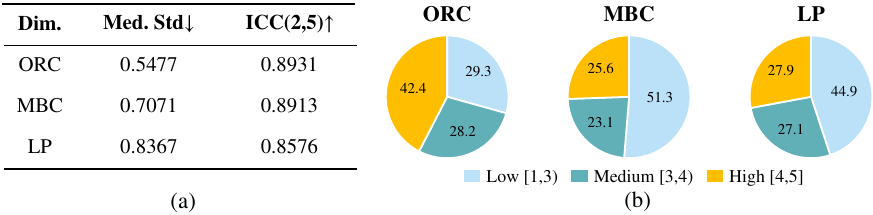}
    \caption{
        Annotation reliability and score distribution. 
(a) Annotation consistency across three dimensions. 
(b) Score distributions for ORC, MBC, and LP.
    }
    \label{fig:rating} 
    \vspace{-5pt} 
\end{figure}
\noindent\textbf{Perception alignment ablation.} To verify the alignment of our scoring model with human judgment, we assessed 1500 generated videos from ten different object removal methods on our VORD benchmark. These videos were divided into three parts and rated by 15 human annotators, with five people in each group responsible for one part. 


Statistical analysis of human annotations (\cref{fig:rating}) confirms their reliability for supervision. (1) Minor Divergence. The median standard deviations (Med. Std) for ORC, MBC, and LP are 0.5477, 0.7071, and 0.8367 ($<$1.0). The rating disagreements are minor and largely confined to adjacent scale levels. (2) High Consistency. The Intraclass Correlation Coefficient ICC(2,5) reaches 0.8931, 0.8913, and 0.8576. These high correlations ($>$0.85) demonstrate a strong consensus among the experts, confirming the averaged labels are highly reliable for supervision. (3) Diverse coverage. The score distribution spans low, medium, and high quality ranges instead of concentrating on easy high-score cases. MBC has the highest low-score proportion, 51.3\%, matching our observation that it is the most challenging dimension. 

Leveraging reliable human ratings, we choose VideoLLaMA3~\cite{zhang2025videollama}, Gemini-3.1-pro and original Qwen3-VL-8B, which support multi-video inputs, to calculate the scores of these videos together with our VOR-MDSM, while ensuring strict input-output consistency. The average scores across the three defined evaluation dimensions were then calculated for all videos. Subsequently, to quantify the correlation, we follow TDVE-Assessor's work~\cite{wang2025tdve} and calculate the Pearson $r$~\cite{pearson1895vii}, Spearman $\rho$~\cite{spearman1987proof}, and Kendall $\tau$~\cite{kendall1938new} correlation coefficients between the human scores and the scores from each of the four models on the three dimensions.
\begin{figure}[t]
    \centering
    \includegraphics[width=\linewidth]{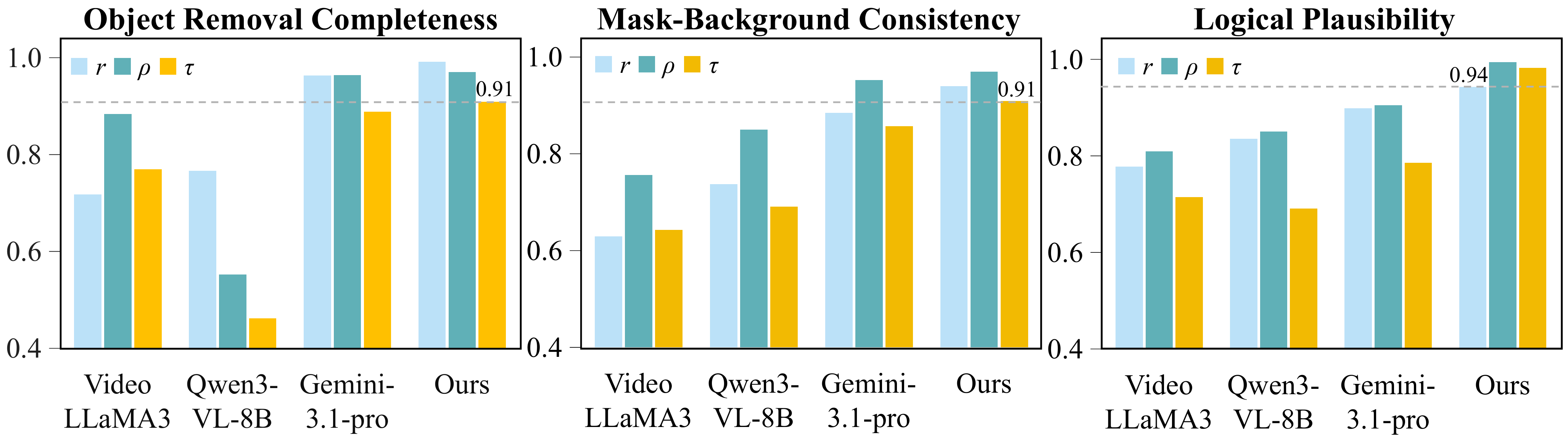}
    \vspace{-8pt}
    \caption{
        Comparative preference consistency of four models across three dimensions. $r$, $\rho$, and $\tau$ denote Pearson, Spearman, and Kendall correlation coefficients, respectively.
    }
    \label{fig:VLM} 
    \vspace{-8pt} 
\end{figure}
The results in \cref{fig:VLM} show that VOR-MDSM achieves $r,\rho,\tau>0.9$ on all three dimensions and outperforms VideoLLaMA3, Qwen3-VL-8B, and Gemini-3.1-pro, demonstrating strong alignment with human ratings. At the method level, the Spearman correlations between human rankings and rankings induced by aggregated traditional metrics, ReMOVE~\cite{chandrasekar2024remove}, and VOR-MDSM are 0.382, 0.697, and 0.976, respectively. Detailed model rankings are provided in the supplementary material (Section 7). 

These findings conclusively demonstrate that VOR-MDSM effectively achieves a human perception-driven evaluation of the removal quality of video objects.
\input{tab/ablation_input_format}
\noindent\textbf{VOR-MDSM training ablation.} Currently, Qwen3-VL supports a multi-video parallel training approach. Meanwhile, inspired by TDVE’s method of temporally concatenating videos, we divide the training input modes into four types: (1) Temporal concatenation: Original video and removal video concatenated along the time dimension to form a single video; (2) Spatial concatenation: Original video, mask video, and removal video are concatenated into a single video from left to right in the spatial dimension; (3) Triple input: Original video, mask video and removal video; (4) Double input: Original video and removal video. This ablation experiment is conducted with strictly consistent parameters across all four types. As shown in \cref{tab:3}, the performance is characterized by calculating the correlation with human ratings. The Alignment Score is the overall mean correlation across all dimensions and metrics, computed as $\frac{1}{3}\sum_{d\in\{\mathrm{ORC},\mathrm{MBC},\mathrm{LP}\}}\frac{r_d+\rho_d+\tau_d}{3}$. Ultimately, the double input that exhibits the best performance is established as the training input for the scoring model.\\
\begin{figure}[!t]
    \centering
    \includegraphics[width=0.8\linewidth]{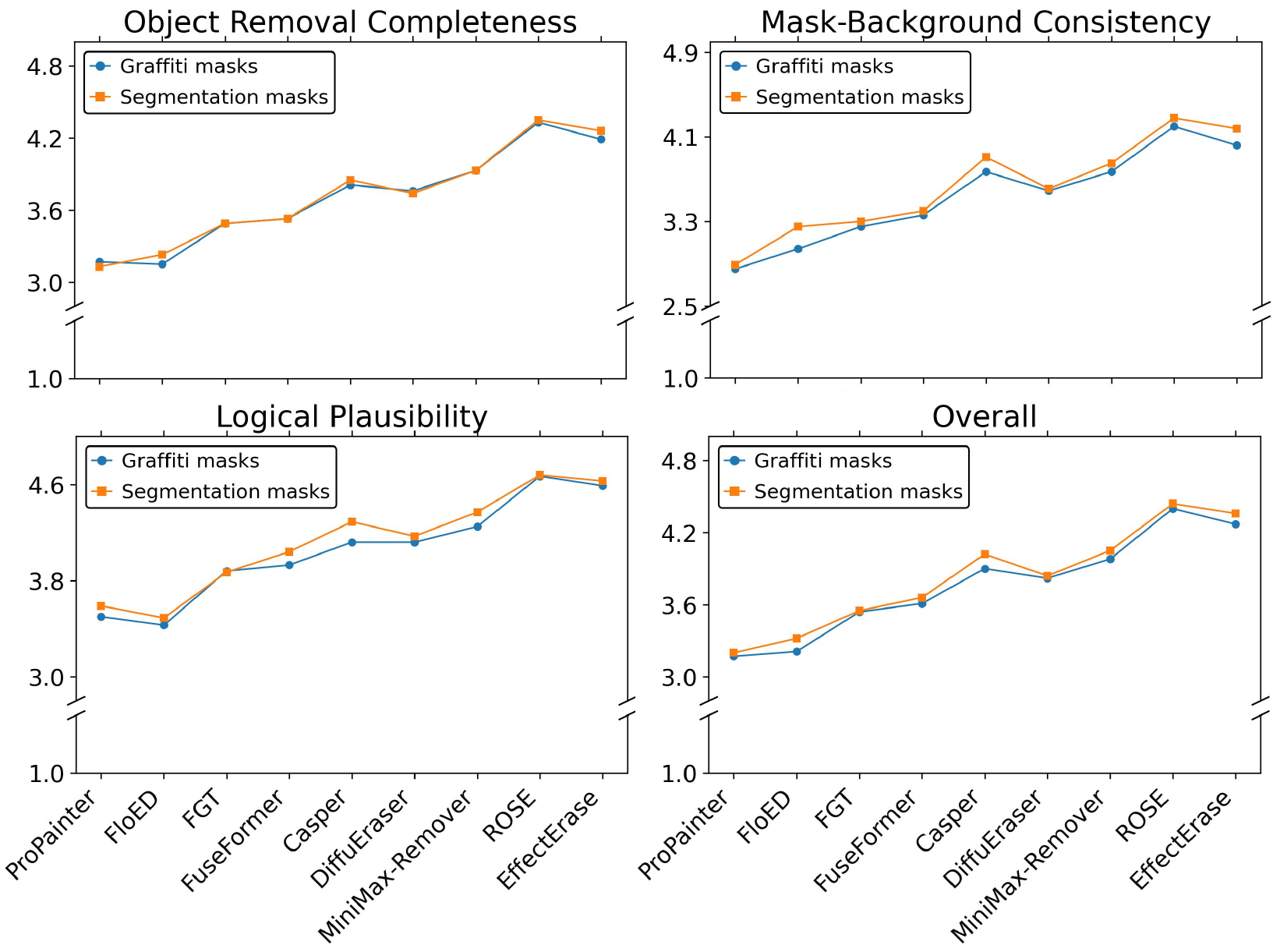}
    \caption{
        Ablation studies on graffiti masks across three dimensions.
    }
    \label{fig:ablation_mask} 
    \vspace{-20pt} 
\end{figure}\noindent\textbf{Dataset scale ablation.} Some existing video object removal models require high inference costs. For instance, Casper takes more than 30 minutes to process a 150-frame video on a single GPU. Therefore, a practical benchmark must strike a careful balance between statistical significance and evaluation efficiency. To rigorously determine the dataset scale and verify if 150 pairs are sufficient to draw stable evaluation conclusions, we conduct an ablation study on the dataset scale. We construct an extended set containing 300 video pairs and intentionally increase the proportion of camera-captured data to 50\% in order to introduce more extreme physical-world diversity. We then re-evaluate all baseline models on this extended set to observe if the capability rankings fluctuate with increased data volume. The evaluation shows an unchanged tier structure from VORD (150 pairs) to VORD-L (300 pairs):
a) \textbf{Tier-1} (score $\geq$ 4.2): \{ROSE, EffectErase, VOID\};
b) \textbf{Tier-2} (3.9 -- 4.2): \{Casper, MiniMax-Remover\};
c) \textbf{Tier-3} (3.6 -- 3.9): \{DiffuEraser, FGT\};
d) \textbf{Tier-4} ($\le$ 3.6): \{FuseFormer, FloED, ProPainter\}.
The model scoring tier is strictly consistent with the \cref{tab:overall}, confirming that 150 pairs already suffice to reliably separate model performance. VORD provides a highly discriminative and unbiased testing platform, while also saving significant inference time and computational resources for the research community.\\
\noindent\textbf{Graffiti masks ablation.} To assess whether existing removal models can segment target objects, we replace the first-frame graffiti masks, which better reflect practical user input, with finely segmented masks and compare the scores of the two resulting output sets (\cref{fig:ablation_mask}). VOID is excluded because its mandatory four-value mask prevents a fair comparison. With finely segmented masks, the models achieve higher mask-background consistency scores, and all except FGT also improve in logical plausibility. In contrast, object removal completeness remains largely unchanged for most models, consistent with the nature of this dimension. Casper changes more than the other models, possibly because it was originally designed for layer separation. These results indicate that target segmentation in current removal models can still be improved.

\begin{wraptable}{r}{0.5\linewidth}
\centering
\scriptsize
\vspace{-4mm}
\begin{tabular}{cccc}
\toprule
    {Dataset} & {Pearson $r$} & {Spearman $\rho$} & {Kendall $\tau$} \\ \midrule
    DAVIS & 0.80 & 0.91 & 0.82 \\
    ROSE-Bench & 0.93 & 0.97 & 0.92 \\
    VORD (Ours) & 0.96 & 0.98 & 0.93 \\
  \bottomrule
\end{tabular}
\vspace{4mm}
\caption{Cross-dataset correlation with human preference.}
\label{tab:cd}
\end{wraptable}
\noindent\textbf{Cross-dataset ablation.} To evaluate the generalization of VOR-MDSM, we conduct cross-dataset validation on the widely used DAVIS and ROSE-Bench datasets under identical experimental settings. We collect human ratings and calculate their correlations with VOR-MDSM scores. As shown in \cref{tab:cd}, the high correlations on both datasets confirm the cross-dataset robustness of VOR-MDSM. In contrast, traditional metrics produce substantially different model rankings across benchmarks, demonstrating their sensitivity to evaluation references. Detailed traditional-metric results and the cross-axis statistics of VOR-MDSM are provided in the supplementary material (Section 7). \\
\noindent\textbf{Inference time.} Under identical full-video inputs, VOR-MDSM averages 0.99s per video, compared with 1.22s for VideoLLaMA3 and 21.50s for the three traditional metrics, demonstrating practical benchmark-scale efficiency.

%% file: tab/ablation_input_format.tex
\begin{table}[t]

\centering
\renewcommand\arraystretch{1.15}

\setlength{\tabcolsep}{5pt} 
\resizebox{\linewidth}{!}{%
\begin{tabular}{@{}ccccccccccc c@{}}
\toprule
 &  &
 \multicolumn{3}{c}{ORC} &
 \multicolumn{3}{c}{MBC} &
 \multicolumn{3}{c}{LP} &
 \\ 
\cmidrule(lr){3-5} \cmidrule(lr){6-8}\cmidrule(lr){9-11}
\multirow{-2}{*}{Model} &
\multirow{-2}{*}{Input Format} &
$r$ & $\rho$ & $\tau$ &
$r$ & $\rho$ & $\tau$ &
$r$ & $\rho$ & $\tau$ &
\multirow{-2}{*}{\shortstack{Alignment\\Scores}} \\
\midrule
 & Temp. concate. & 0.1612 & 0.0958 & 0.0364 & 0.4340 & 0.1905 & 0.1429 & 0.4438 & 0.0952 & 0.0714 & 0.1857 \\
 & Spat. concate.  & 0.9691 & 0.9758 & 0.9436 & 0.8683 & 0.8796 & 0.7412 & 0.8680 & 0.9157 & 0.8154 & 0.8863 \\
 & Triple input      & 0.7338 & 0.6627 & 0.5185 & 0.4891 & 0.5270 & 0.3273 & 0.6306 & 0.6190 & 0.4286 & 0.5485 \\
\multirow{-4}{*}{Ori} &
 Double input        & 0.7668 & 0.5522 & 0.4619 & 0.7376 & 0.8503 & 0.6910 & 0.8354 & 0.8503 & 0.6910 & 0.7152 \\
\midrule
 & Temp. concate. & 0.9789 & 0.9940 & 0.9820 & 0.9244 & 0.9286 & 0.7857 & 0.9146 & 0.9524 & 0.8571 & 0.9242 \\
 & Spat. concate.  & 0.9885 & 0.9940 & 0.9820 & 0.9134 & 0.9048 & 0.7857 & 0.9033 & 0.9048 & 0.7857 & 0.9069 \\
 & Triple input      & 0.9903 & 0.9940 & 0.9820 & 0.9652 & 0.9701 & 0.9092 & 0.8938 & 0.9581 & 0.9092 & 0.9524 \\
\multirow{-4}{*}{Trained} &
 \textbf{Double input} & 0.9919 & 0.9701 & 0.9092 & 0.9399 & 0.9701 & 0.9092 & 0.9429 & 0.9940 & 0.9820 & \textbf{0.9566} \\
\bottomrule
\end{tabular}%
}
\caption{Ablation studies on the input format. $r$: Pearson $r$; $\rho$: Spearman $\rho$;  $\tau$: Kendall $\tau$.}
\label{tab:3}
\end{table}

%% file: sec/5_conclusion.tex
\section{Conclusion}
\label{sec:conclusion}
This work introduces VOR-Bench to evaluate video removal models. VOR-Bench overcomes the limitations of existing VOR evaluation sets and metrics, providing more credible evaluation results. The proposed VOR dataset, VORD, with paired edited videos, diverse data types and graffiti masks, better aligns with practical application scenarios. Our novel data framework, rMPAF, provides a viable solution for generating paired edited videos in various scenarios. In addition, we present a specialized human perception-driven multi-modal evaluator for VOR, VOR-MDSM. This evaluator assesses three key dimensions: object removal completeness, mask-background consistency and logical plausibility. Overall, VOR-Bench provides references for the performance improvement of video removal models. 

%% file: sec/X_suppl.tex
\clearpage
\setcounter{page}{1}
\begin{center}
{\LARGE \bfseries Supplementary Material}
\end{center}
\begin{figure}[!b]
    \centering
    \includegraphics[width=0.73\linewidth]{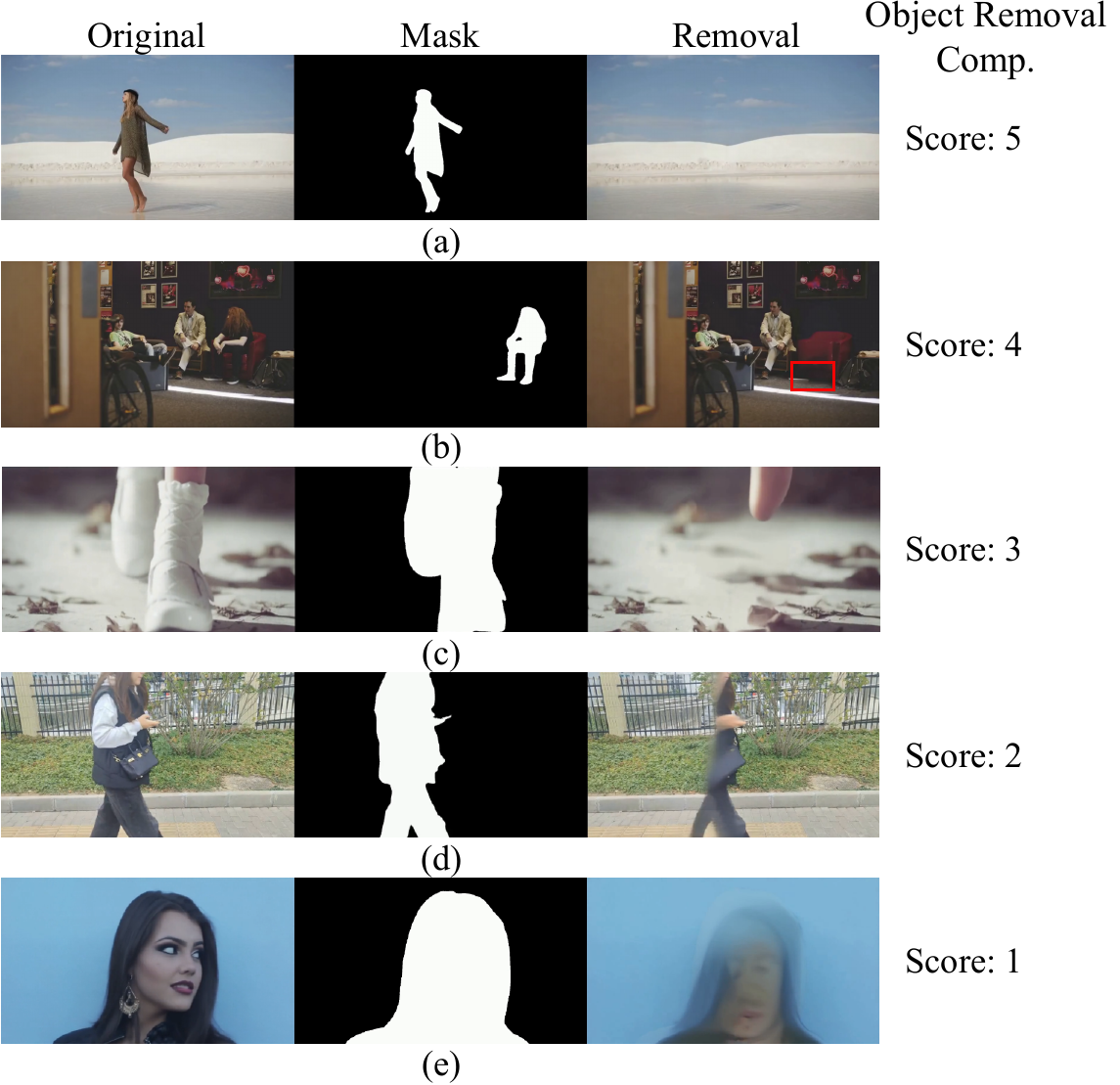}
    \caption{
         Visual examples of the object removal completeness. Zoom in to observe a more pronounced effect.
    }
    \label{fig:orc} 
\end{figure}
\section{More Details of Scoring Information}
\label{sec:score}
\subsection{Detailed Information of Vision Experts}
There are 15 visual experts, each of whom has practical experience in video editing and possesses a professional background in the field. They have participated in subjective evaluation experiments for text-to-image and text-to-video models before, gaining a thorough understanding of aspects such as image quality, aesthetics, distortion in the spatial domain, and issues like multi-frame artifacts and frame-to-frame deformation consistency in the temporal domain. For this work, they all underwent thorough training prior to the evaluation.

\subsection{Detailed Information of Scoring Criteria}
\textbf{Object removal completeness.} The scoring rules are as follows:

Score 1: object removal completion degree 0\%$\sim$50\%; 

Score 2: object removal completion degree 50\%$\sim$70\%;

Score 3: object removal completion degree 70\%$\sim$90\%;

Score 4: object removal completion degree 90\%$\sim$100\%;

Score 5: object removal completion degree 100\%.

Specific examples are shown in \cref{fig:orc}.
\begin{figure}[t]
    \centering
    \includegraphics[width=0.73\linewidth]{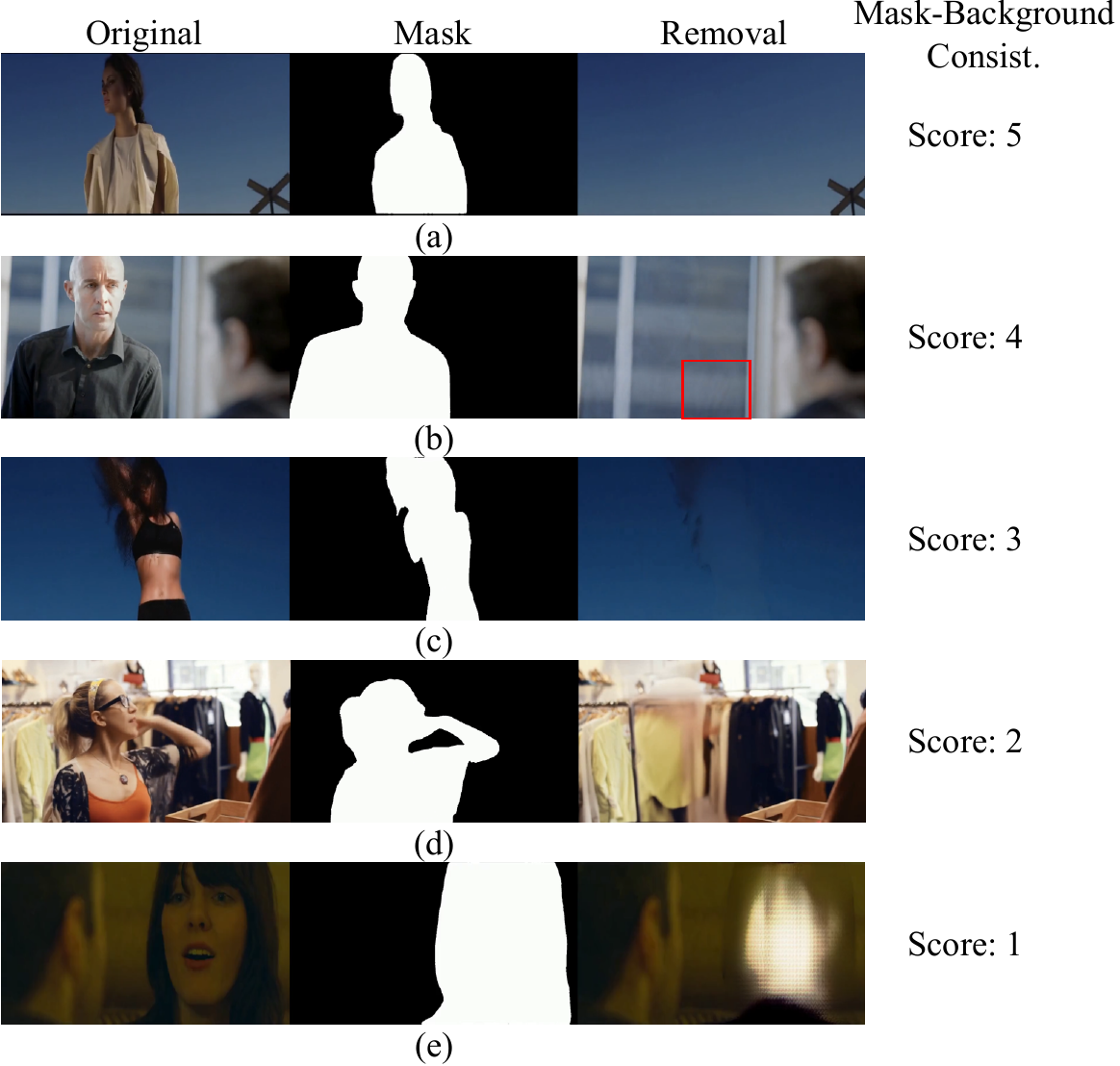}
    \caption{
         Visual examples of the mask-background consistency. Zoom in to observe a more pronounced effect.
    }
    \label{fig:mbc} 
\end{figure}
\begin{figure}[!t]
    \centering
    \includegraphics[width=0.73\linewidth]{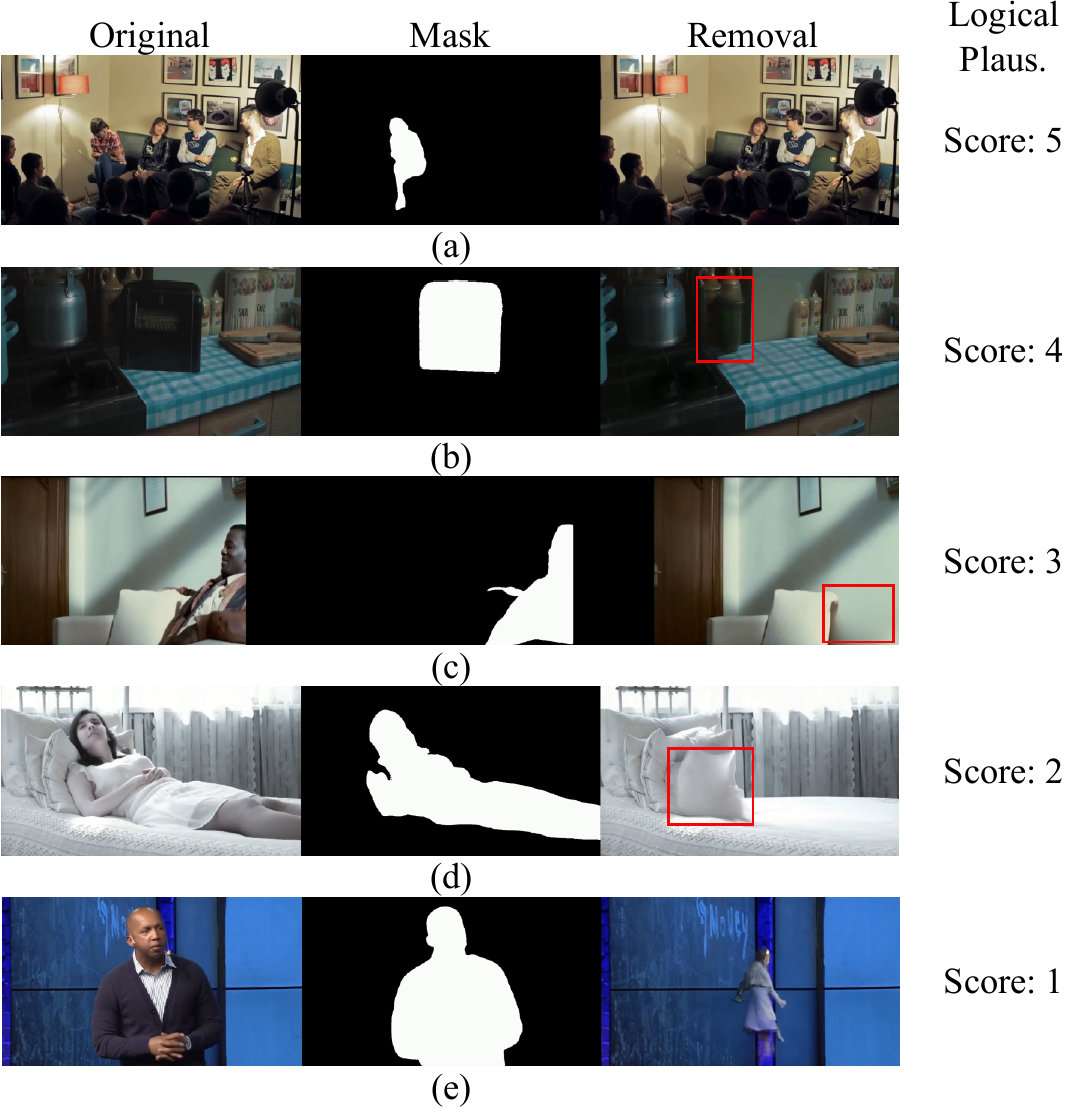}
    \caption{
         Visual examples of the logical plausibility. Zoom in to observe a more pronounced effect.
    }
    \label{fig:lp} 
\end{figure}

\noindent\textbf{Mask-background consistency.}
The scoring rules are as follows: 

Score 1: the removed area does not blend with the background, and there are large areas of holes, artifacts, and color differences in the removed area, which affect the visual quality; 

Score 2: the transition between the removed area and the background is poor, with obvious boundaries and a sense of disharmony; 

Score 3: moderate fusion, with occasional visible boundaries or seams, and inconsistent textures or styles; 

Score 4: the removed area blends well with the background, with slight differences in boundaries or textures;

Score 5: the removed area blends perfectly with the background.

Specific examples are shown in \cref{fig:mbc}.\\\textbf{Logical plausibility.}
The scoring rules are as follows: 

Score 1: the generation of hallucinated objects;

Score 2: obvious logical deviations, such as deformed shapes and moving still objects;

Score 3: related generation to the background, but there are slight logical issues; 

Score 4: more subtle logical issues can only be seen by zooming in on the video;

Score 5: No issues with any physical laws or scene logic.

Specific examples are shown in \cref{fig:lp}. In addition, the sample (d) in \cref{fig:lp} shows the unreasonable movement of a static pillow, and please refer to the specific video for more intuitive display.

All the examples above are accompanied by corresponding videos.

\section{Additional Quantitative Results}
\label{sec:additional_results}

\subsection{Metric-Induced Model Rankings}

Since traditional metrics, ReMOVE, and VOR-MDSM operate on different score scales, we compare their induced model rankings with the human preference ranking. As shown in \cref{tab:supp_rank}, VOR-MDSM achieves a substantially higher Spearman correlation with the human ranking than traditional metrics and ReMOVE, further demonstrating its alignment with human perception.

\begin{table}[!t]
\centering
\scriptsize
\setlength{\tabcolsep}{1.8pt}
\renewcommand{\arraystretch}{1.0}
\resizebox{\linewidth}{!}{%
\begin{tabular}{@{}lcccccccccc@{}}
\toprule
& FuseF. & FGT & ProP. & FloED & Casp. & DiffE. & MiniM. & ROSE & EffE. & VOID \\
\midrule
ReMOVE$\uparrow$
& 0.884 & \underline{0.901} & 0.886 & 0.841 & 0.887
& 0.892 & 0.897 & \textbf{0.902} & 0.898 & 0.895 \\

ReMOVE rank
& 9 & 2 & 8 & 10 & 7 & 6 & 4 & 1 & 3 & 5 \\

Trad. rank$^{\dagger}$
& 9 & 10 & 2 & 8 & 5 & 4 & 6 & 3 & 7 & 1 \\

VOR-MDSM rank
& 8 & 7 & 9 & 10 & 4 & 6 & 5 & 1 & 2 & 3 \\

Human rank
& 7 & 8 & 9 & 10 & 5 & 6 & 4 & 1 & 2 & 3 \\
\midrule

\multicolumn{11}{c}{
Spearman $\rho$ w.r.t.\ human ranking:
Trad.\ $0.382$;\quad
ReMOVE $0.697$;\quad
\textbf{VOR-MDSM $0.976$}
} \\
\bottomrule
\end{tabular}%
}
\vspace{0.5mm}
{\footnotesize
$^{\dagger}$The traditional ranking aggregates PSNR, SSIM, and LPIPS.
FuseF.: FuseFormer; ProP.: ProPainter; Casp.: Casper; DiffE.: DiffuEraser;
MiniM.: MiniMax-Remover; EffE.: EffectErase.
}
\caption{Comparison of metric-induced model rankings on VORD.}
\label{tab:supp_rank}
\end{table}

\subsection{Traditional Metrics Across Benchmarks}

We further evaluate traditional metrics on DAVIS and ROSE-Bench. As shown in \cref{tab:supp_bench}, the resulting model rankings vary substantially across benchmarks, indicating their sensitivity to evaluation references. The VORD column additionally reports the mean and standard deviation across the three VOR-MDSM dimensions. The relatively small standard deviations indicate balanced cross-axis evaluation.

\begin{table}[!t]
\centering
\normalsize
\setlength{\tabcolsep}{3pt}
\renewcommand{\arraystretch}{1.05}

\begin{tabular}{@{}lccc|ccc|c@{}}
\toprule
& \multicolumn{3}{c|}{DAVIS}
& \multicolumn{3}{c|}{ROSE-Bench}
& VORD \\
Model
& PSNR$\uparrow$
& SSIM$\uparrow$
& LPIPS$\downarrow$
& PSNR$\uparrow$
& SSIM$\uparrow$
& LPIPS$\downarrow$
& Avg$\pm$Std \\
\midrule

FuseFormer
& \underline{33.937}
& \underline{0.960}
& 0.033
& 28.673
& 0.921
& 0.065
& $3.393\pm0.222$ \\

FGT
& 32.405
& 0.943
& 0.042
& 28.810
& 0.920
& 0.067
& $3.669\pm0.278$ \\

ProPainter
& \textbf{36.674}
& \textbf{0.980}
& \textbf{0.015}
& 29.428
& 0.930
& 0.058
& $3.380\pm0.228$ \\

FloED
& 21.742
& 0.656
& 0.110
& 27.728
& 0.910
& 0.070
& $3.327\pm0.199$ \\

Casper
& 27.803
& 0.862
& 0.082
& 29.725
& 0.923
& 0.063
& $4.149\pm0.164$ \\

DiffuEraser
& 33.307
& 0.958
& \underline{0.032}
& 28.944
& 0.925
& 0.062
& $3.887\pm0.224$ \\

MiniMax-Remover
& 29.798
& 0.888
& 0.059
& 28.565
& 0.923
& 0.065
& $4.091\pm0.198$ \\

ROSE
& 27.401
& 0.866
& 0.071
& \textbf{32.799}
& \textbf{0.939}
& \underline{0.052}
& $\mathbf{4.327\pm0.208}$ \\

EffectErase
& 22.912
& 0.773
& 0.095
& 30.607
& \underline{0.937}
& 0.056
& $\underline{4.269\pm0.238}$ \\

VOID
& 28.029
& 0.880
& 0.063
& \underline{31.393}
& 0.933
& \textbf{0.038}
& $4.218\pm0.251$ \\

\bottomrule
\end{tabular}
\vspace{4mm}
\caption{Traditional metrics on DAVIS and ROSE-Bench, together with the cross-axis statistics of VOR-MDSM on VORD. Bold and underlined values denote the best and second-best results, respectively.}
\label{tab:supp_bench}
\end{table}

\begin{figure}[!t]
    \centering
    \includegraphics[width=\linewidth]{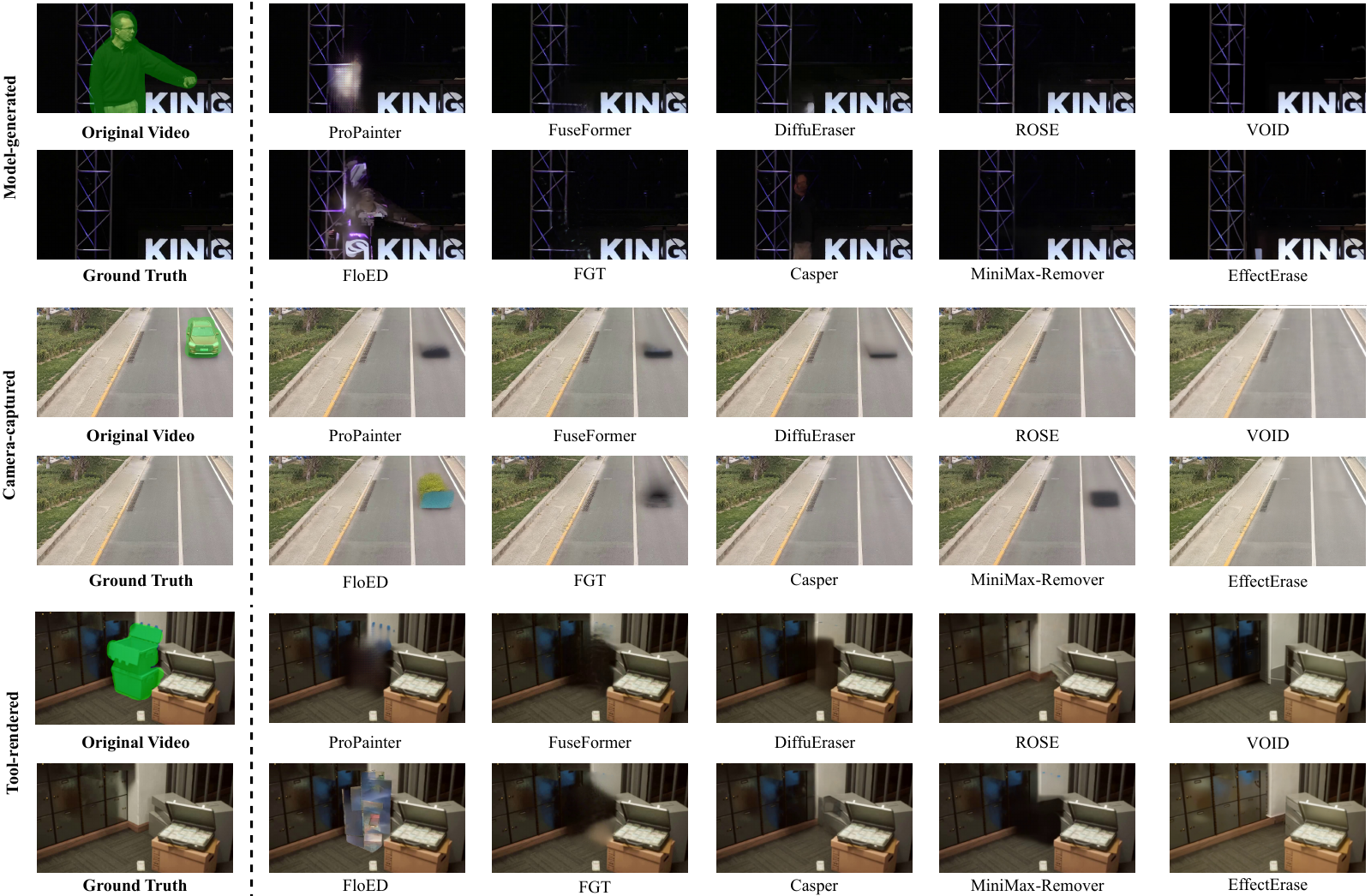}
    \caption{
         The visual results of different models across three data types: model-generated data, camera-captured data and tool-rendered data.
    }
    \label{fig:visual_x} 
\end{figure}
\section{More Visualization Samples}
\label{sec:Experiment}
More visualization samples are shown in \cref{fig:visual_x}. We have also provided the corresponding video files, which can be found in the attachments as \path{tool-rendered-visual-samples}, \path{model-generated-visual-samples} and \path{camera-captured-visual-samples}.